\documentclass{article}

\usepackage{iclr2025_conference,times}

\usepackage{url}
\definecolor{citecolor}{RGB}{17,80,197}
\definecolor{linkcolor}{HTML}{ED1C24}
\usepackage[breaklinks=true,colorlinks,citecolor=citecolor,linkcolor=linkcolor,bookmarks=false]{hyperref}

\usepackage[utf8]{inputenc} % allow utf-8 input
\usepackage[T1]{fontenc}    % use 8-bit T1 fonts
\usepackage{booktabs}       % professional-quality tables
\usepackage{amsfonts}       % blackboard math symbols
\usepackage{nicefrac}       % compact symbols for 1/2, etc.
\usepackage{microtype}      % microtypography
\usepackage{xcolor}         % colors
\usepackage{enumitem}

\usepackage{graphicx}
\usepackage{pifont}
\usepackage{amsmath,amsthm,amssymb,bm,stmaryrd,bbm,mathtools}
\usepackage{xspace}
\usepackage{multirow} 
\usepackage{wrapfig}
\usepackage[normalem]{ulem}
\useunder{\uline}{\ul}{}
\usepackage{float}

\newcommand{\ours}{\textsc{SiDR}\xspace}
\newcommand{\oursf}{$\textsc{SiDR}_{\text{full}}$\xspace}

\newcommand{\oursb}{$\textsc{SiDR}_{\beta}$\xspace}

\newcommand{\dpr}{\textsc{DPR}\xspace}
\newcommand{\vdr}{\textsc{VDR}\xspace}
\newcommand{\vdrb}{$\textsc{VDR}_{\beta}$\xspace}

\newcommand{\innerproduct}[2]{\langle #1, #2 \rangle}

\title{Semi-Parametric Retrieval via Binary Bag-of-Tokens Index}

\author{Jiawei Zhou$^{1,3}$ 
    \quad Li Dong$^2$ 
    \quad Furu Wei$^2$ 
    \quad Lei Chen$^{1,3}$ \\
The Hong Kong University of Science and Technology$^1$ \qquad
Microsoft Research$^2$ \\
The Hong Kong University of Science and Technology (Guangzhou)$^3$ \\
\texttt{\{jzhoubu,leichen\}@ust.hk}, 
\texttt{\{lidong1,fuwei\}@microsoft.com}}

\iclrfinalcopy
\begin{document}
\maketitle

\begin{abstract}
Information retrieval has transitioned from standalone systems into essential components across broader applications, with indexing efficiency, cost-effectiveness, and freshness becoming increasingly critical yet often overlooked.
In this paper, we introduce \textbf{S}em\textbf{I}-parametric \textbf{D}isentangled \textbf{R}etrieval (\ours), a bi-encoder retrieval framework that decouples retrieval index from neural parameters to enable efficient, low-cost, and parameter-agnostic indexing for emerging use cases. 
Specifically, in addition to using embeddings as indexes like existing neural retrieval methods, \ours supports a non-parametric bag-of-tokens index for search, achieving BM25-like indexing complexity with significantly better effectiveness. 
Our comprehensive evaluation across 16 retrieval benchmarks demonstrates that \ours outperforms both neural and term-based retrieval baselines under the same indexing workload:
(i)~When using an parametric embedding-based index, \ours exceeds the performance of conventional neural retrievers while maintaining similar training complexity;
(ii)~When using a non-parametric tokenization-based index, \ours matches the complexity of traditional term-based retrieval BM25, while consistently outperforming it on in-domain datasets;
(iii)~Additionally, we introduce a late parametric mechanism that matches BM25 index preparation time for search while outperforming both BM25 and other neural retrieval baselines in effectiveness. Code is available at \url{https://github.com/jzhoubu/sidr}.
\end{abstract}

\section{Introduction}
In recent years, information retrievers has evolved from end-to-end systems to essential components in various applications, including question answering~\citep{kolomiyets2011survey,zhu2021retrieving}, classification~\citep{long2022retrieval}, recommendation~\citep{dong2020mamo,manzoor2022towards} and dialog systems~\citep{liu2024chatqa}. This evolution has notably accelerated with the advent of the retrieval-augmented generation (RAG) paradigm~\citep{bommasani2021opportunities,guu2020retrieval,yu2022survey,mialon2023augmented}, in which the retrieval component enables large language models (LLMs) to access relevant data from external sources, effectively addressing challenges such as like hallucination~\citep{ji2023survey,zhang2023siren}, obsolescence~\citep{wang2023survey}, and privacy concerns~\citep{huang2022large}. 

Traditional retrieval systems provide end-to-end search services and build indexes offline, without concern for cost or latency. In contrast, advanced retrieval components integrate with various downstream models, requiring greater flexibility to meet the diverse needs of applications. We present several \textbf{emerging retrieval scenarios} where current neural retrievers face limitations, and introduce specific \textbf{\textit{index properties}} in our proposed framework designed to overcome these challenges.

\begin{wrapfigure}{!t}{0.45\textwidth}
  \centering
  \vspace{-1pt}
  \includegraphics[width=0.45\textwidth]{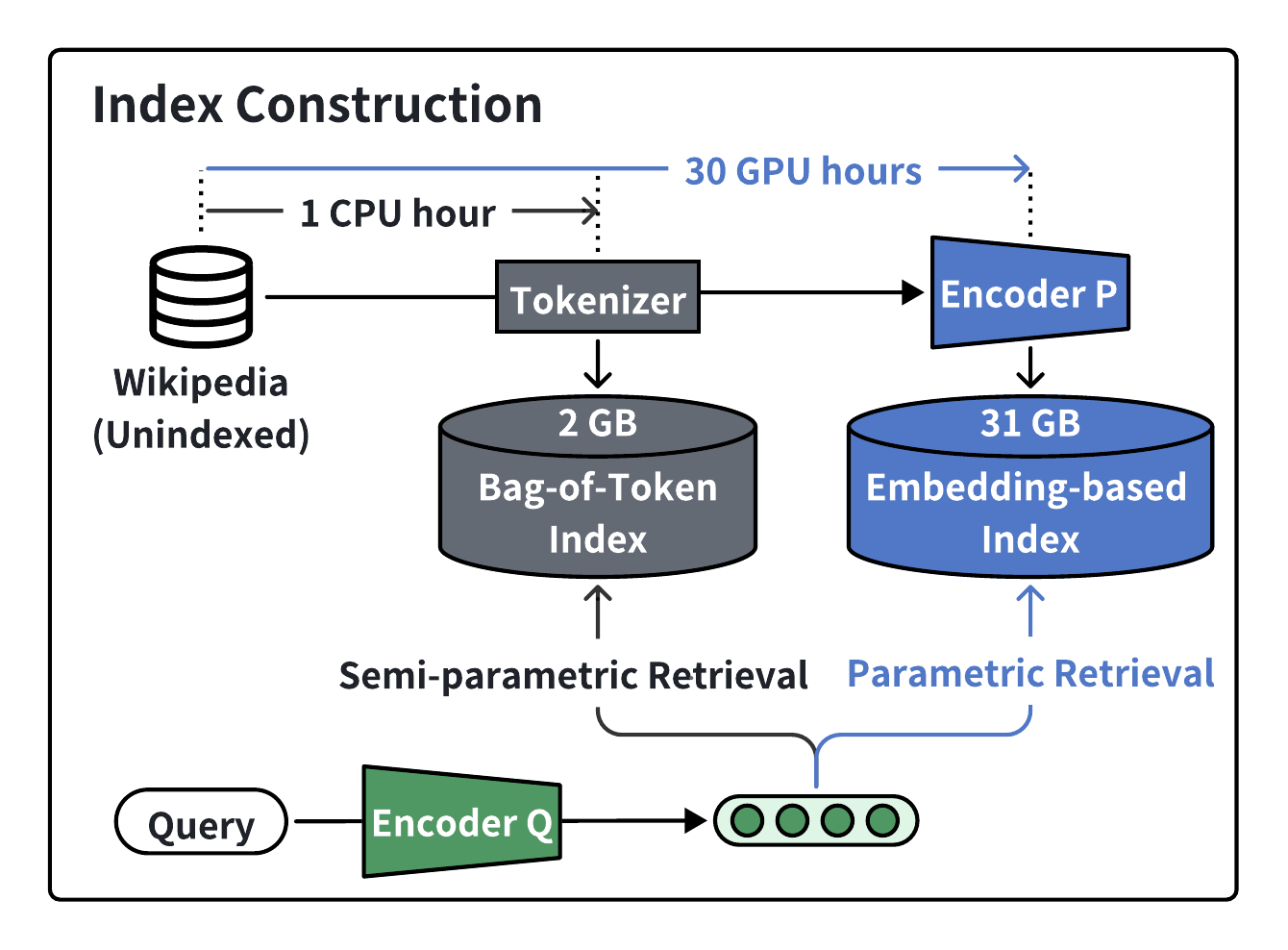}
  \vspace{-18pt}
  \caption{Comparison of storage (2 GB vs. 31 GB) and resource costs (1 CPU hour vs. 30 GPU hours) between two indexes.}
  \label{fig:fig1}
  \vspace{-18pt}
\end{wrapfigure}

\paragraph{Scenario 1: Online indexing for RAG applications with real-time knowledge sources.} 
In RAG applications that depend on real-time knowledge source, such as up-to-the-minute internet information~\citep{liu2023fingpt} and user-uploaded content requiring immediate responses~\citep{wang2024survey}, efficient \textbf{\textit{online indexing}} is essential as it determines the time lag between the availability of data and its application. Effective neural retrieval with efficient online indexing facilitates the rapid assimilation and filtering of real-time datastreams, reducing computational burdens and mitigating hallucinations~\citep{liu2024lost,shuster2021retrieval} when LLMs process long contexts. 

\paragraph{Scenario 2: Low-cost index for exploration and deployment.} Concerns about data privacy~\citep{huang2022large, arora2023knowledge} and licensing issues~\citep{min2023silo} are driving startups and individual developers towards building their own local retrieval pipelines for RAG applications. In this context, it is common to construct temporary retrieval indexes that are frequently modified and rebuilt to cater to different data sources and to optimize for varying chunk size selections, allowing for effective analysis of large datastores~\citep{shao2024scaling} and the fine-tuning of configurations. 
Given their sensitivity to resource constraints, smaller entities often prioritize cost reduction over maximizing effectiveness, making \textbf{\textit{Low-cost Index}} more important than achieving state-of-the-art performance.

\paragraph{Scenario 3: Parameter-agnostic index for co-training retrievers with LLMs.} 
A significant challenge in co-training neural retrievers with LLMs is the index update issue~\citep{asai2023retrieval} caused by in-training retrieval~\citep{guu2020retrieval,izacard2022atlas,shi2023replug}. Specifically, during training, a neural retriever parameterized by $\theta$ is learned to search information from a datastore $\mathcal{D}$ to enhance the LLMs. The retrieval index, denoted as $E_{\theta}(\mathcal{D})$, consists of the neural embeddings of the datastore $\mathcal{D}$. As the parameters update $\theta\rightarrow\theta'$, the datastore index needs to be rebuilt $E_{\theta}(\mathcal{D}) \rightarrow E_{\theta'}(\mathcal{D})$ to prevent it from becoming stale. This process is computationally expensive and compromises the training objectives. Developing a neural retrieval that supports a \textbf{\textit{Parameter-agnostic Index}} could address this issue and streamline the co-training pipelines.

To meet these emerging needs, our paper introduces the \textbf{S}em\textbf{I}-parametric \textbf{D}isentangled \textbf{R}etrieval framework (\ours), which decouples retrieval index from neural parameters to facilitate an \textit{efficient}, \textit{low-cost}, and \textit{non-parametric} indexing setup. Specifically, our framework involves learning parametric term weighting within a language model vocabulary space, where non-parametric representations can be straightforwardly defined and constructed via tokenization. By aligning these two types of representations, one encoder within the bi-encoder framework can optionally utilize tokenization-based representations as a shortcut for indexing large data volumes. As a result, \ours simultaneously supports a parametric index that utilizes neural embeddings and a non-parametric index that employs bag-of-tokens representations. As illustrated in Figure~\ref{fig:fig1}, using the non-parametric index for the Wikipedia corpus drastically reduces the indexing cost and time from 30 GPU hours to just 1 CPU hour and reduces storage size from 31GB to 2GB. This design offers flexibility in choosing indexes with varying complexity to meet diverse retrieval scenarios and co-training propose.

Our comprehensive evaluations across 16 retrieval benchmarks demonstrate that \ours outperforms both neural and term-based retrieval baselines under comparable indexing workloads. Specifically, our framework with a non-parametric index achieves a 10.6\% improvement in top-1 accuracy in-domain compared to BM25, while maintaining indexing efficiency on par with BM25. Additionally, when utilizing a parametric index, our framework surpasses neural retrieval methods by 2.7\% with similar training complexity. Furthermore, our late parametric approach that retrieves from a non-parametric index and re-ranks the results on-the-fly, achieving indexing efficiency comparable to BM25 while maintaining the effectiveness of neural retrieval.

We summarize our contributions from two main aspects:
\begin{itemize}[leftmargin=*]
    \item From IR perspective: We introduce a versatile semi-parametric retrieval framework that supports both parametric and non-parametric indexes to accommodate diverse downstream scenarios. The parametric index uses neural embeddings for effectiveness, while the non-parametric relies solely on tokenization for efficiency. We further propose a late parametric mechanism to maximize the trade-off between retrieval effectiveness and indexing efficiency.
    \item From RAG perspective: Our approach introduces in-training retrieval on a fixed non-parametric index, which avoids index staleness and eliminates the need for costly index rebuilds within the retriever's training loop. This simplifies the co-training the retrieval system with other models.
\end{itemize}

\section{Background}

\paragraph{Information retrieval task.} 
Given a query $q$ and a datastore $\mathcal{D}$, information retrieval~\citep{manning2009introduction} aims to identify the most relevant passage $p \in \mathcal{D}$ based on $q$. This task is typically performed using a bi-encoder framework, which employs two independent encoders to embed queries and passages into vector representations. The retrieval process can be formulated as:
\begin{gather*}
\hat{p} = {\mathrm{argmax}}\, f(q, \mathcal{D}) = \underset{\forall p \in \mathcal{D}}{\mathrm{argmax}}\, f(q, p) \nonumber
% f(q, p) = \innerproduct{E(q)}{E(p)} \nonumber
\end{gather*}
In this equation, $\hat{p}$ is the retrieved passage, and $f$ is a function measures the relevance between $q$ and $p$, usually calculated as the inner product of their vector representations.

\paragraph{Notation.}
We use $E_{\theta}(\cdot)$ to denote a general neural embedding process, applicable to both dense and sparse retrieval.
Specifically, for term-based and sparse lexical retrieval, a $|V|$-dimensional representation is used, where each dimension represents the weight of a token or word within vocabulary $V$. We denote this embedding function as $V_{\text{Model}_{\theta}}(\cdot): x \rightarrow \mathcal{R}^{|V|}$, where the subscript indicates the model architecture, and $\theta$ reflects whether the embedding involves learnable parameters.

\paragraph{Traditional term-based retrieval.}
Traditional term-based retrieval, such as TF-IDF~\citep{ramos2003using} and BM25~\citep{robertson2009probabilistic}, assess relevance based on weighted term overlap, which can be described as:
\begin{align}
f_{\text{BM25}}(q, p) 
= \innerproduct{V_{\text{BM25}}(q)}{V_{\text{BM25}}(p)} 
= \innerproduct{w_{\text{BM25}} \cdot V_{\text{BoW}}(q)}{w_{\text{BM25}} \cdot V_{\text{BoW}}(p)} \nonumber
\end{align}
These methods do not involve learned neural parameters and are therefore categorized as non-parametric~\citep{min2022nonparametric,freeman2002example}. They employ a million-scale dimensional bag-of-words (BoW) representation $V_{\text{BoW}}(\cdot)$, with heuristic statistical metrics determining the term weighting $w_{\text{BM25}}$ for each dimension. 
Due to the efficiency and cost-effectiveness in constructing the term-based index $V_{\text{BM25}}(\mathcal{D})$, these methods are still widely used in industry applications.

\paragraph{Neural retrieval.}
Unlike term-based retrieval that is heuristic-driven, neural retrieval~\citep{karpukhin2020dense,zhu2023large} is data-driven and parameterized, tailored to learn on specific datasets and tasks. The relevance assessment is defined as:
\begin{align}
f_{\theta}(q, p) &= \innerproduct{E_{\theta}(q)}{E_{\theta}(p)} \nonumber
\end{align}
While neural retrievers are effective with ample training data, the construction of the parametric index $E_{\theta}(\mathcal{D})$ requires embedding the entire datastore, which introduces significant computational costs and latency that hinder their widespread adoption.

Neural retrievers can be further categorized into entangled retrievers (a.k.a. dense retrievers) and disentangled retrievers (a.k.a. sparse lexical retrievers), categories which we detail in Appendix~\ref{sec:appendix-texonomy}. Specifically, entangled retrievers utilize latent representations with dimensions such as 768, 1024, or 2048. In contrast, disentangled retrievers employ disentangled representations, with dimensions equal to vocabulary sizes, such as 30522 for BERT's vocabulary, where each dimension reflects the importance of each token.
\section{Methodology}
As an overview, \textbf{S}em\textbf{i}-parametric \textbf{D}isentangled \textbf{R}etrieval (\ours) is a disentangled retrieval system that builds on the \vdr architecture~\citep{zhou2024retrievalbased}, incorporating modifications in the learning objective and utilizing in-training retrieval for negative mining to enhance its effectiveness. 
At downstream, \ours supports both embedding-based index and bag-of-tokens index. The primary goal of our work is to expand the functionality of the current retrieval system to better accommodate emerging scenarios. 

In the following sections, we delve into details of representation (§\ref{sec:representation}), rational (§\ref{sec:mlm}), training objectives (§\ref{sec:objective}), search pipelines (§\ref{sec:search}), and the application of in-training retrieval techniques (§\ref{sec:negative}).

\begin{figure*}[ht]
\begin{center}
\includegraphics[width=0.99\textwidth]{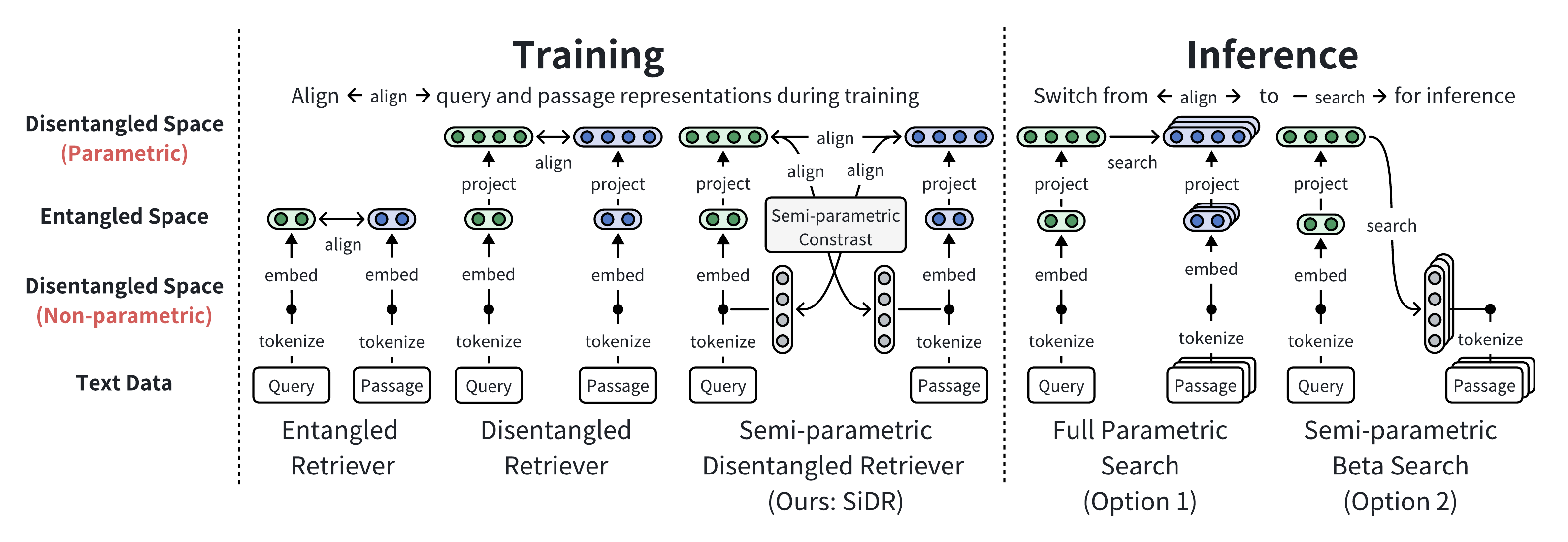}
\caption{Left: Training frameworks of entangled retriever, disentangled retriever and our proposed semi-parametric disentangled retriever \ours; Right: Different inference pipelines of \ours.}
\label{fig:framework}
\end{center}
\vspace{-0.4cm}
\end{figure*}

\subsection{Parametric and Non-parametric Representation}
\label{sec:representation}
Disentangled retrievers typically represent data in a language model vocabulary space, which can be interpreted as a set of tokens with weights. Our parametric representation, $V_{\theta}(x)$, uses token weights learned by a neural encoder, whereas the non-parametric representation, $V_{\text{BoT}}(x)$, can be viewed as using unweighted tokens generated by a tokenizer.

\paragraph{Parametric representation.} We inherit the \vdr encoder~\citep{zhou2024retrievalbased}, which extends the conventional MLMs architecture with three modification: (i) replacing the softmax activation with $elu1p$ to map dimensional values from $(0,1)$ to $(0, +\infty)$; (ii) applying max-pooling to aggregate token representations into a global representation; and (iii) employing top-$k$ sparsify ($\text{S}_{topk}$) to prune the less significant dimensional values to zero. These modifications can be expressed as follows:
\begin{equation}\label{eq:elu1p}
  elu1p(x) =
  \begin{cases}
    x+1 & \text{if $x >= 0$} \\
    e^{x} & \text{otherwise} \nonumber
  \end{cases}
\end{equation}
\begin{equation}\label{eq:vdr}
  V_{\theta}(x) = \text{S}_{topk} \circ \text{MaxPool} \circ \{V_{\text{MLMH}_{\theta}+\text{elu1p}}(t_i|x), \forall t_i \in x \} \nonumber
\end{equation}

Significantly, these modifications aggregate token representations into a global one, while preserving the property of assigning larger values to more relevant dimensions.

\paragraph{Non-parametric representation.} The non-parametric bag-of-tokens (BoT) representation for a sequence of tokens $x$ is defined as follows: 
\begin{align}
 V_\text{BoT}(x) = \text{MaxPool} \circ \{V_\text{BoT}(t_i), \forall t_i \in x\} \nonumber;\quad
 V_\text{BoT}(x)[i] = 
    \begin{cases}
        1 & \text{if $V[i]\in x$} \\
        0 & \text{otherwise} \nonumber
  \end{cases}   
\end{align}
$V_\text{BoT}(x)$ can be seen as the result of applying max pooling to the one-hot representations of all tokens in $x$, assigning each $i$-th dimension a value of 1 or 0, depending on whether the $i^{th}$ token in $V$ is present in $x$. Compared to $V_\text{BM25}(\cdot)$, $V_\text{BoT}(\cdot)$ is tokenizer-specific, with dimensionality on the scale of tens of thousands, and uses binary values that require less storage space, making it well-suited for tensorization and efficient GPU computation.

\subsection{Rationale of semi-parametric alignment.} 
\label{sec:mlm}
In this section, we elaborate on the rationale behind the semi-parametric design. The semi-parametric alignment is designed to be consistent with mask language models (MLMs)~\citep{devlin2018bert} pre-training objecitve.

\paragraph{Mask language model pre-training.}
During pre-training, MLMs are optimized to predict masked tokens by leveraging the context. Specifically, given an input sequence of tokens $x = [t_1, t_2, \dots, M(t_i), \dots, t_n]$ with $t_i$ masked, the MLM uses its prediction head (MLMH) with a softmax function to produce the probability $V_{\text{MLMH}_\theta + \text{softmax}}(\text{Mask}(t_i)|x)$ of the masked token $t_i$ over a vocabulary. The ground truth probability is the one-hot representation of the token $t_i$. In this paper, we refer to this type of representation as the bag-of-tokens (BoT) representation, denote as $V_{\text{BoT}}(t_i)$. The mask token prediction task can be viewed as alignment between the vocabulary distribution of the masked token position with the one-hot representation $V_{\text{BoT}}(t_i)$ in a masked setup:
\begin{equation}\label{eq:mlm}
 V_{\text{MLMH}_{\theta}+\text{softmax}}(\text{Mask}(t_i)|x) \xleftrightarrow{\text{align}}
 V_{\text{BoT}}(t_i)
\end{equation}
As a result, the representation $V_{\text{MLMH}_{\theta}}(t_{i}|x)$ tends to assign large values to the dimension corresponding to $t_{i}$, or that are semantically related to $t_{i}$ based on the context $x$. 

\paragraph{Semi-parametric alignment.}
The alignment between parametric and non-parametric representation can be expressed as follows:
\begin{equation}\label{eq:semi-align}
  \text{S}_{topk} \circ \text{MaxPool} \circ \{V_{\text{MLMH}_{\theta}+\text{elu1p}}(t_i|x), \forall t_i \in x \} \xleftrightarrow{\text{align}} 
  \text{S}_{topk} \circ  \text{MaxPool} \circ \{V_\text{BoT}(t_i), \forall t_i \in x \} \nonumber
\end{equation}
The semi-parametric alignment is modeled after the MLM pre-training objective, as detailed in Equation~\ref{eq:mlm}, and expands by aligning multi-token representations between the query and passage. The consistency between upstream pre-training and downstream tuning supports the alignability of these two representations. 

\subsection{Semi-parametric Contrastive Learning}
\label{sec:objective}
In a batch containing $N$ instances, each instance consists of a query $q_i$, a positive passage $p_i$, and a set of of negative passages. 
Our training objective is based on contrastive learning~\citep{jaiswal2020survey}, which aims to maximize the similarity of positive pairs $f(q_i, p_i)$ for all instances $i$, while minimize the similarity of all negative pairs, denoted as $f(q_i, p_j)$ for all $j \neq i$. 
The loss function is defined as follows:
\begin{small}
\begin{equation}
\begin{split}
L(q, p) = &-\sum_{i=1}^{N}(
\log{
\underbrace{
    \frac{e^{{f(q_i,p_i)}}}
        {\sum_{\forall p \in B} e^{f(q_i,p)}}}
_\text{q-to-p}}
+ 
\log{
\underbrace{
    \frac{e^{f(p_i,q_i)}}{\sum_{\forall q \in B} e^{f(p_i,q)}}}
_\text{p-to-q}}) \nonumber
\end{split}
\end{equation}
\end{small}
This results in a final loss that integrates both parametric and semi-parametric components:
\begin{small}
\begin{equation}
\begin{split}
L_{\text{para}}(q,p) &= L(V_{\theta}(q), V_\theta(p))
\\
L_{\text{semi-para}}(q,p) &= L(V_\theta(q), V_{\text{BoT}}(p))/2 + L(V_{\text{BoT}}(q), V_\theta(p))/2
\\
L_{\text{final}}(q,p) &= L_{\text{para}}(q,p) + L_{\text{semi-para}}(q,p)
\nonumber
\end{split}
\end{equation}
\end{small}

The parametric contrastive loss $L_{p}$ aims to align the parametric representations of $q$ and $p$, a common objective for retrieval training. The semi-parametric contrastive loss $L_{sp}$ ensures interaction between the non-parametric and parametric representations, which forms the foundation of our model to support a BoT index $V_\text{BoT}(\mathcal{D})$.

\subsection{Search Pipelines and Index Types}
\label{sec:search}
Our framework supports both a parametric embedding-based index and a non-parametric tokenization-based index. Below, we discuss search functions and the corresponding index type.

\noindent \textbf{Full parametric search} (\oursf) utilizes a \textbf{parametric index} $V_{\theta}(\mathcal{D})$, which relies on embeddings derived from a neural encoder for the datastore. The relevance is defined as: \begin{equation} f_{\theta}(q, \mathcal{D}) = \innerproduct{V_{\theta}(q)}{V_{\theta}(\mathcal{D})} \nonumber \end{equation} 
This is the common indexing process for neural retrieval systems, which are effective but involve higher costs and longer latency for embedding the entire $\mathcal{D}$ to obtain the index $V_{\theta}(\mathcal{D})$.

\noindent \textbf{Semi-parametric beta search} (\oursb) leverages a \textbf{non-parametric index} $V_{\text{BoT}}(\mathcal{D})$ based on BoT representations of the datastore, which are constructed solely by a tokenizer. The relevance is defined as: 
\begin{equation}
f_{\beta}(q, \mathcal{D}) = \innerproduct{V_{\theta}(q)}{V_{\text{BoT}}(\mathcal{D})} \nonumber 
\end{equation} 

Beta search applies BoT representations on the index side, eliminating the need for neural embeddings during index processing for large datastores, making it suitable for various applications.

\noindent \textbf{Late parametric with top-m re-rank} (\oursb($m$)) is a search pipeline that starts search with a non-parametric index to retrieve top-$m$ passages, denote as $\mathcal{D}_{m}$, and then embeds them for re-ranking: 
\begin{align}
f_{\beta}(q, \mathcal{D}) = \innerproduct{V_{\theta}(q)}{V_{\text{BoT}}(\mathcal{D})} \nonumber; \quad
f_{\theta}(q, \mathcal{D}_{m}) = \innerproduct{V_{\theta}(q)}{V_{\theta}(\mathcal{D}_{m})} \nonumber
\end{align}

Late parametric retrieval requires a first-stage retriever to support a non-parametric index, followed by a second-stage bi-encoder retriever to re-rank and cache the passage embeddings. From an efficiency and cost perspective, the late parametric with top-$m$ re-rank only requires embedding at most $N_{q} \times m$ passages, where $N_{q}$ is the number of queries. It starts the search with a non-parametric index, distributing the embedding workload throughout the online search process, achieving a fast retrieval setup. When dealing with exploratory scenarios that have limited queries or extremely large datastores, this approach becomes more efficient, as $N_{q} \times m$ remains much smaller than $|\mathcal{D}|$.

\subsection{In-training Retrieval for Negative Sampling}
\label{sec:negative}
While semi-parametric retrieval offers advantages for in-training retrieval by eliminating the need for re-indexing, it also has drawbacks, notably its limited effectiveness due to using non-parametric representations on index side. To enhance their performance, we integrate beta search in the training loop to dynamically source hard negative passages, leveraging the strengths of semi-parametric design to counterbalance their limitations. 
Specifically, during the training, \ours employs beta search to retrieve the top-$m$ passages in real-time --- using $V_{\theta}(q)$ to search on non-parametric index $V_{\text{BoT}}(\mathcal{D})$ and get the top-$m$ results $\mathcal{D}_{m}$. 
Subsequently, each passage in $\mathcal{D}_{m}$ is assessed whether it is negative based on exact matches with the answer strings. For each query, one negative is randomly selected from the identified negatives. This method is exclusively used for the Wikipedia benchmark, which provides answer strings to distinguish between negative and positive passages. 

While in-training retrieval is increasingly adopted to enhance retrieval training~\citep{zhan2021optimizing,xiong2020approximate} and facilitate co-training retrievers with LLMs~\citep{shi2023replug}, previous approaches have necessitated periodic index refreshment in the training loop. In contrast, our approach uniquely leverages a fixed index $V_{\text{BoT}}(\mathcal{D})$, eliminating the need for re-indexing $\mathcal{D}$. Our analysis shows that incorporating in-training retrieval with our non-parametric index does not add significant latency; however, a slight latency increase occurs due to the string matching process to identify negatives, which is discussed in Section~\ref{sec:implementation}.
\section{Experimental Setup}
\label{sec:setup}

\subsection{Datasets}

\paragraph{Wiki21m benchmark.} 
Following established benchmark in retrieval literature~\citep{chen2017reading,karpukhin2020dense}, we train our model on the training splits of Natural Questions (NQ;~\citeauthor{kwiatkowski2019natural}, \citeyear{kwiatkowski2019natural}), TriviaQA (TQA;~\citeauthor{joshi2017triviaqa}, \citeyear{joshi2017triviaqa}), and WebQuestions (WQ;~\citeauthor{berant2013semantic}, \citeyear{berant2013semantic}) datasets, and evaluated it on their respective test splits. The retrieval corpus used is Wikipedia, which contains over 21 million 100-word passages.

\paragraph{BEIR benchmark.} 
We train our model on MS MARCO passage ranking dataset~\citep{bajaj2016ms}, which consists of approximately 8.8 million passages with around 500 thousand queries. The performance is assessed both in-domain on MS MARCO and in a zero-shot setting across 12 diverse datasets within the BEIR benchmark \citep{thakur2021beir}.

\subsection{Baselines}
\paragraph{Primary baselines.} 
We primarily compare our model with several established retrieval baselines, selected due to their similar model sizes and training complexities, ensuring a comparable training cost. These include dense retrieval \dpr, term-based retrieval BM25, and sparse lexical retrieval models such as SPLADE~\citep{formal2021splade}, uniCOIL~\citep{lin2021few}, and \vdr~\citep{zhou2024retrievalbased}. Specifically, \vdrb refer to directly using the \vdr models to perform beta search.

\paragraph{Advanced baselines.} 
We also introduce advanced retrieval systems for baselines, including ANCE~\citep{xiong2020approximate}, LexMAE~\citep{shen2022lexmae}, SPLADE-v2~\citep{formal2021splade_v2}, Contriever~\citep{izacard2021unsupervised}, GTR~\citep{ni2021large}, E5~\citep{wang2022text}, and Dragon~\citep{lin2023train}.
These systems leverage larger foundational models~\citep{ma2023fine,ni2021large}, retrieval-oriented pre-training~\citep{fan2022pre,zhou2022hyperlink}, or knowledge distillation~\citep{formal2022distillation} to enhance performance. 
We have categorized them as advanced baselines due to their significantly higher training costs associated with the additional training techniques. Future work may explore their integration with our model to assess potential benefits.

\subsection{Implementation Details}
\label{sec:implementation}
\paragraph{Hyperparameters.}
For the NQ, TQA, and WQ datasets, our model is trained for 80 epochs, utilizing in-training retrieval for negative sampling. For the MS MARCO dataset, the training duration is set to 40 epochs. We utilize a batch size of 128 and an AdamW optimizer~\citep{loshchilov2018decoupled} with a learning rate set at $2 \times 10^{-5}$. Our model use a top-$k$ sparsification with $k=768$, matching the dimensionality of conventional dense retrieval embeddings. For computational devices, our systems are equipped with 4 NVIDIA A100 GPUs and Intel Xeon Platinum 8358 CPUs.

\paragraph{Training cost.} 
The training durations for \dpr, \vdr, and \ours on the NQ dataset are 5, 8, and 9 hours respectively, with 80 epochs under similar conditions. For \ours, the training time per epoch without in-training retrieval is 6 minutes, which is identical to \vdr. This duration increases to 8 minutes per epoch when incorporating in-training retrieval. The additional time is primarily due to the string matching process required to identify negative samples from the retrieved top-$k$ passages.
\section{Experiments}
\label{sec:exp}
\subsection{Main Results}

\begin{table*}[ht]
\centering
\caption{Top-1/5/20 retrieval accuracy on test sets (i.e., percentage of questions for which the answers is found in the retrieved passages). \textbf{Bold} numbers indicate the best performance within each setting.}
\scalebox{0.8}{
\setlength{\tabcolsep}{2mm}{
\begin{tabular}{lccccccccc}
\hline
\addlinespace[0.5ex]
 & \multicolumn{3}{c}{NQ} & \multicolumn{3}{c}{TQA} & \multicolumn{3}{c}{WQ} \\ 
\cmidrule(lr){2-4} \cmidrule(lr){5-7} \cmidrule(lr){8-10} 
 & top1 & top5 & top20 & top1 & top5 & top20 & top1 & top5 & top20 \\ \hline
 \multicolumn{9}{l}{\textit{Parametric Index}} \\
%\quad $\text{E5-base}$ & 58.0 & 78.4 & 86.4 & 58.7 & 74.6 & 82.2 & 46.9 & 68.7 & 75.0 \\
%\quad $\text{Contriever}$ & 41.5 & 66.8 & 74.2 &  &  & & &  & \\
\quad $\text{DPR}$ & 46.0 & 68.9 & 80.2 & 54.1 & 71.5 & 80.0 & 37.4 & 59.7 & \textbf{73.2} \\
\quad $\text{VDR}$ & 43.8 & 68.0 & 79.9 & 52.9 & 71.3 & 79.3 & 37.1 & 58.7 & 72.5 \\
% \quad $\text{ANCE}$ & - & \textbf{70.7} & \textbf{81.4} & - & \textbf{73.9} & \textbf{81.4} & - & \textbf{65.7} & \textbf{77.2} \\
\quad \oursf & \textbf{49.1} & \textbf{69.3} & \textbf{80.7} & \textbf{56.2} & \textbf{73.0} & \textbf{80.5} & \textbf{40.2} & \textbf{61.0} & \textbf{73.2} \\ \hline
 \multicolumn{9}{l}{\textit{Non-parametric Index}} \\
\quad BM25 & 22.7 & 43.6 & 62.9 & 48.2 & 66.4 & 76.4 & 19.5 & 42.6 & 62.8 \\
\quad $\text{VDR}_{\beta}$ & 12.3 & 30.0 & 46.8 & 16.9 & 31.6 & 45.9 & 7.7 & 22.4 & 39.2 \\
\quad \oursb & \textbf{39.8} & \textbf{62.9} & \textbf{76.3} & \textbf{50.4} & \textbf{70.7} & \textbf{79.5} & \textbf{32.1} & \textbf{54.1} & \textbf{69.8} \\ \hline
 \multicolumn{9}{l}{\textit{Late parametric with top-$m$ re-rank}} \\
\quad $\text{BM25+DPR}$~$(m=5)$ & 32.2 & 43.6 & 62.9 & 54.8 & 66.4 & 76.4 & 28.0 & 42.6 & 62.8 \\
\quad $\text{BM25+DPR}$~$(m=20)$ & 39.4 & 55.5 & 62.9 & 55.4 & 71.0 & 76.4 & 34.6 & 53.2 & 62.8 \\
\quad $\text{BM25+DPR}$~$(m=100)$ & 44.4 & 63.6 & 73.5 & 56.6 & 72.3 & 80.5 & 39.9 & 59.2 & 70.2 \\
\quad \oursb~$(m=5)$ & 44.9 & 60.6 & 76.4 & 54.9 & 70.7 & 79.6 & 38.0 & 54.1 & 69.8 \\
\quad \oursb~$(m=20)$ & 49.5 & 68.2 & 76.4 & 56.7 & 72.9 & 79.5 & 39.6 & 59.5 & 69.8 \\
\quad \oursb~$(m=100)$ & \textbf{50.3} & \textbf{70.7} & \textbf{80.6} & \textbf{56.8} & \textbf{73.3} & \textbf{81.3} & \textbf{41.5} & \textbf{62.0} & \textbf{73.5} \\ \hline
\end{tabular}
}
}
\label{table:results_wiki21m}
\end{table*}

\paragraph{Wiki21m benchmark results.}  
As shown in Table~\ref{table:results_wiki21m}, when using a parametric index, \oursf outperforms \dpr and \vdr in top-1 retrieval accuracy by 2.6\% and 3.8\%, respectively. These results demonstrate that although our primary objective is to enable neural retrieval with support for non-parametric indexing, our modifications do not diminish effectiveness with an embedding-based index; in fact, they may even improve it. This enhancement also suggests that current benchmarks may favor lexical relevance, likely due to their construction relying on term-based retrieval methods.

When utilizing a non-parametric index, \oursb significantly surpasses \vdrb and BM25 in top-1 accuracy by 28.5\% and 10.6\%, respectively. Unlike BM25, which relies on empirically derived heuristic term weights for indexing, \oursb employs binary values in a significantly smaller vocabulary space, facilitating tensorization on GPUs. With ample in-domain training data, \oursb significantly outperforms BM25, demonstrating the exceptional learnability and generalizability of neural retrievers in this context. 

Additionally, late parametric methods improve \oursb by enabling on-the-fly re-ranking of the top-$m$ passages. Our results show that by re-ranking the top-20 passages, \oursb~$(m=20)$ matches the performance of \oursf, and by extending re-ranking to the top-100 passages, it surpasses all primary baselines. 
Beyond effectiveness, the most significant advantage of late parametric methods is their substantial reduction in computational costs. For example, in evaluations on the NQ test split with 3k queries, \oursf requires embedding the entire $\mathcal{D}$, which consists of 21 million passages. In contrast, \oursb~$(m=100)$ only needs to embed $N_{q} \times m$ passages, amounting to just 1\% of the passages in $\mathcal{D}$, yet achieves superior effectiveness. 
This underscores the exceptional suitability of late parametric methods for exploration or evaluation scenarios.

\paragraph{BEIR benchmark results.}
As shown in Table~\ref{table:beir-effectiveness}, \oursf surpasses $\text{VDR}$, $\text{DPR}$, and other primary baselines in the BEIR benchmark when using either a parametric index or a non-parametric index with late parametric techniques, consistent with the findings from the Wiki21m benchmark. However, when relying solely on a non-parametric index, \oursb outperforms $\text{VDR}_\beta$ and BM25 on in-domain datasets but falls behind BM25 on most out-of-domain datasets.
We attribute this performance decline to three factors. First, due to the lack of answer strings to accurately identify negative passages in MS MARCO, we do not implement in-training retrieval during training on MS MARCO, which likely contributes to weaker performance. Second, as non-parametric indexes lack neural parameters, they are more sensitive to shifts in data distribution, which may lead to weaker effectiveness in out-of-domain scenarios. Lastly, many BEIR datasets exhibit a lexical bias due to their construction using BM25, as noted in the BEIR paper~\citep{thakur2021beir}, which inherently gives BM25 an advantage.

\begin{table*}[ht]
\centering
\caption{Retrieval performance on MS MARCO (MRR@10) and BEIR benchmark (NDCG@10). \textbf{Bold} numbers indicate the best performance within each setting.}
\scalebox{0.7}{
\setlength{\tabcolsep}{1.1mm}{
\begin{tabular}{lcccccccccccccc}
\hline
\addlinespace[0.5ex]
  & \rotatebox{70}{MS MARCO} & \rotatebox{70}{ArguAna} & \rotatebox{70}{Climate-FEVER} & \rotatebox{70}{DBPedia} & \rotatebox{70}{FEVER} & \rotatebox{70}{FiQA} & \rotatebox{70}{HotpotQA} & \rotatebox{70}{NFCorpus} & \rotatebox{70}{NQ} & \rotatebox{70}{SCIDOCs} & \rotatebox{70}{SciFact} & \rotatebox{70}{TREC-COVID} & \rotatebox{70}{Touché-2020} & Avg.  \\ \hline
 \multicolumn{15}{l}{\textit{Advanced Retrieval Baselines}} \\
\quad $\text{LexMAE}$ & \textbf{48.0} & 50.0 & 21.9 & 42.4 & \textbf{80.0} & 35.2 & \textbf{71.6} & 34.7 & 56.2 & 15.9 & 71.7 & 76.3 & \textbf{29.0} & \textbf{48.7} \\
\quad $\text{Splade}$-$\text{v2}$ & 43.3 & 47.9 & 23.5 & \textbf{43.5} & 78.6 & 33.6 & 68.4 & 33.4 & 52.1 & 15.8 & 69.3 & 71.0 & 27.2 & 47.0 \\
\quad $\text{Contriever}$ & - & 44.6 & 23.7 & 41.3 & 75.8 & 32.9 & 63.8 & 32.8 & 49.8 & 16.5 & 67.7 & 59.6 & 23.0 & 44.3 \\
\quad $\text{GTR}$-$\text{base}$ & 42.0 & 51.1 & \textbf{24.1} & 34.7 & 66.0 & 34.9 & 53.5 & 30.8 & 49.5 & 14.9 & 60.0 & 53.9 & 20.5 & 41.2 \\
\quad $\text{E5}$-$\text{base}$ & 43.1 & \textbf{51.4} & 15.4 & 41.0 & 58.2 & \textbf{36.4} & 63.3 & \textbf{36.1} & \textbf{62.9} & \textbf{19.0} & \textbf{73.1} & \textbf{79.6} & 28.3 & 47.1 \\
\quad $\text{Dragon}$ & 39.3 & 48.9 & 22.2 & 41.7 & 78.1 & 35.6 & 64.8 & 32.9 & 53.1 & 15.4 & 67.5 & 74.0 & 24.9 & 46.6 \\ 
\quad $\text{ANCE}$ & 33.8 & 41.5 & 19.8 & 28.1 & 66.9 & 29.5 & 45.6 & 23.7 & 44.6 & 12.2 & 50.7 & 65.4 & 28.4 & 38.0 \\ \hline
 \multicolumn{15}{l}{\textit{Parametric Index}} \\
\quad $\text{DPR}$ & 30.2 & 40.8 & 16.2 & 30.4 & 63.8 & 23.7 & 45.2 & 26.1 & 43.2 & 10.9 & 47.4 & 60.1 & 22.1 & 35.8 \\
\quad $\text{UniCOIL}$ & 32.9 & 35.5 & 15.0 & 30.2 & 72.3 & 27.0 & 64.0 & 32.5 & 36.2 & 13.9 & \textbf{67.4} & 59.7 & 25.9 & 39.4 \\
\quad $\text{SPLADE}$ & 34.0 & 43.9 & \textbf{19.9} & 36.6 & 73.0 & 28.7 & 63.6 & 31.3 & 46.9 & 14.5 & 62.8 & 67.3 & 20.1 & 42.4 \\
\quad $\text{VDR}$ & \textbf{34.3} & 48.6 & 17.6 & 39.0 & \textbf{74.0} & 28.8 & \textbf{65.5} & \textbf{33.0} & 47.2 & \textbf{15.3} & 67.3 & 67.8 & \textbf{29.8} & 44.5 \\
\quad \oursf & 34.2 & \textbf{53.0} & 17.9 & \textbf{39.3} & 71.5 & \textbf{29.8} & 65.4 & \textbf{33.0} & \textbf{47.7} & 15.1 & 66.2 & \textbf{68.0} & 29.7 & \textbf{44.7} \\ \hline
 \multicolumn{15}{l}{\textit{Non-parametric Index}} \\
\quad $\text{BM25}$ & 18.7 & 31.5 & \textbf{21.3} & \textbf{31.3} & \textbf{75.3} & \textbf{23.6} & \textbf{60.3} & \textbf{32.5} & \textbf{32.9} & \textbf{15.8} & \textbf{66.5} & \textbf{65.6} & \textbf{36.7} & \textbf{41.1} \\
\quad $\text{VDR}_{\beta}$ & 6.1 & 14.1 & 6.1 & 7.9 & 28.4 & 6.4 & 5.7 & 23.9 & 6.8 & 8.1 & 54.5 & 21.9 & 9.2 & 16.1 \\
\quad \oursb & \textbf{19.0} & \textbf{38.6} & 10.8 & 20.8 & 46.5 & 19.8 & 49.4 & 27.9 & 25.3 & 11.1 & 64.2 & 53.5 & 23.7 & 32.5 \\ \hline
 \multicolumn{15}{l}{\textit{Late parametric with top-$m$ re-rank}} \\
\quad \oursb~$(m=10)$ & 26.3 & 44.0 & 12.1 & 24.9 & 57.3 & 22.8 & 55.4 & 30.2 & 33.1 & 12.2 & 65.5 & 54.9 & 25.2 & 36.5 \\
\quad \oursb~$(m=20)$ & 29.2 & 48.0 & 13.8 & 30.6 & 62.3 & 25.7 & 58.7 & 32.3 & 38.4 & 13.8 & 65.6 & 59.1 & 27.3 & 39.6 \\
\quad \oursb~$(m=100)$ & \textbf{32.9} & \textbf{51.5} & \textbf{16.6} & \textbf{37.8} & \textbf{69.2} & \textbf{29.3} & \textbf{63.3} & \textbf{33.2} & \textbf{44.7} & \textbf{14.8} & \textbf{65.7} & \textbf{65.9} & \textbf{29.0} & \textbf{43.4} \\ \hline

% \quad BM25+E5~$(m=100)$ & \textbf{32.9} & \textbf{51.5} & \textbf{16.6} & \textbf{37.8} & \textbf{69.2} & \textbf{29.3} & \textbf{63.3} & \textbf{33.2} & \textbf{44.7} & \textbf{14.8} & \textbf{65.7} & \textbf{65.9} & \textbf{29.0} & \textbf{43.4} \\ \hline
\end{tabular}
}
}
\label{table:beir-effectiveness}
\end{table*}

\subsection{Retrieval Latency}
We evaluated the latency of various retrieval systems across different stages using NQ test split and Wikipedia corpus, as shown in Table~\ref{table:main-efficiency}. The comparison assumes that both BM25 and SiDR indexes fit entirely into CPU/GPU memory. Further details can be found in Appendix~\ref{sec:appendix-efficiency}.

\iffalse
\begin{table*}[t]
\centering
\caption{\textcolor{blue}{Latency at each stage of the retrieval pipeline. $\text{T}(\cdot)$: computations of tokenization; $\text{E}_{\theta}(\cdot)$: computation of neural model forward. ${\dag}$: Expected latency.}}
\scalebox{0.6}{
\begin{tabular}{lcccccc}
\toprule
\multirow{2}[2]{*}{Model} & \multicolumn{2}{c}{Indexing} & \multicolumn{3}{c}{Search} & \multirow{2}[2]{*}{Total} \\
   & $\text{T}(\mathcal{D})$ & $\text{E}_{\theta}(\mathcal{D})$ & $\text{E}_{\theta}(q)$ & $\text{f}(q,\mathcal{D})$ & $\text{E}_{\theta}(p)$  &  \\
\midrule
\multicolumn{7}{l}{\textit{\textbf{w/o GPU (CPU-only)}}} \\
\quad $\text{BM25}$ & 0.6h & /  & /  & 2m & /  & 0.6h \\
\quad $\text{DPR}$ & /  & 194.4h$^{\dag}$ & 2m & 2m & /  & 194.4h$^{\dag}$ \\
\quad \oursf & /  & 227.0$^{\dag}$ & 2m & 20m & /  & 227.3h$^{\dag}$ \\
\quad \oursb & 0.5h & /  & 2m & 3m & /  & 0.6h \\
\quad \oursb($m$=100) & 0.5h & /  & 2m & 3m & 3.2h & 3.8h \\
\midrule
\textit{\textbf{w/ GPU}} &    &    &    &    &    &  \\
\quad $\text{BM25}$ & 0.6h & /  & /  & 1m & /  & 0.6h \\
\quad $\text{DPR}$ & /  & 20.3h & 12s & 41ms & /  & 20.3h \\
\quad $\text{VDR}$ & /  & 23.7h & 15s & 130ms & /  & 23.7h \\
\quad \oursf & /  & 23.7h & 15s & 130ms & /  & 23.7h \\
\quad \oursb & 0.5h & /  & 15s & 30ms & /  & 0.6h \\
\quad \oursb($m$=100) & 0.5h & /  & 15s & 30ms & 20m & 0.9h \\
\bottomrule
\end{tabular}%

}
\label{table:main-efficiency}
\end{table*}
\fi

\begin{table*}[t]
\centering
\caption{Latency at each stage of the retrieval pipeline. $\text{T}(\cdot)$: computations of tokenization; $\text{E}_{\theta}(\cdot)$: computation of neural model forward. ${\dag}$: Computations performed on a single GPU; times in parentheses indicate the latency if performed on a single CPU thread}. $\text{E}_{\theta}(p)$ in search stage refers to the passage embedding used for late parametric re-ranking.
\scalebox{0.8}{
\begin{tabular}{l|cc|ccccc|c}
\hline
\multirow{2}{*}{Model} & \multicolumn{2}{c|}{Indexing} & \multicolumn{5}{c|}{Search} & \multirow{2}{*}{Total} \\
 & $\text{T}(\mathcal{D})$ & $\text{E}_{\theta}(\mathcal{D})$ & $\text{E}_{\theta}(q)$ & $\text{T}(q)$ & $\text{f}(q,\mathcal{D})$ & $\text{E}_{\theta}(p)$ & Total &  \\ \hline
$\text{BM25}$ & 0.6h & / & / & \multicolumn{2}{c}{40s$^{\dag}$ (2m)} & / & 40s & 0.6h \\
$\text{DPR}$ & / & 20.3h$^{\dag}$ & 12s$^{\dag}$(2m) & / & 41ms$^{\dag}$(2m) & / & 12s & 20.3h \\
$\text{VDR}$ & / & 23.7h$^{\dag}$ & 15s$^{\dag}$(2m) & / & 130ms$^{\dag}$(20m) & / & 15s & 23.7h \\
\oursf & / & 23.7h$^{\dag}$ & 15s$^{\dag}$(2m) & / & 130ms$^{\dag}$(20m) & / & 15s & 23.7h \\
\oursb & 0.5h & / & 15s$^{\dag}$(2m) & / & 30ms$^{\dag}$ (3m) & / & 15s & 0.5h \\
\oursb$(m=20)$ & 0.5h & / & 15s$^{\dag}$(2m) & / & 30ms$^{\dag}$(3m) & 4m$^{\dag}$ & 4m & 0.6h \\
\oursb$(m=100)$ & 0.5h & / & 15s$^{\dag}$(2m) & / & 30ms$^{\dag}$(3m) & 20m$^{\dag}$ & 20m & 0.9h \\ \hline
\end{tabular}
}
\label{table:main-efficiency}
\end{table*}

\vspace{-0.5cm}
\paragraph{Indexing stage.}
The indexing stage converts the textual corpus into a searchable format. Both \oursb and BM25 use tokenzation-based index and can complete indexing within 1 hour on a CPU, much faster than the over 20 hours required on GPUs for embedding-based index. The indexing stage often accounts for a large portion of the overall time and cost in the retrieval pipeline. Our BoT index is efficient, more cost-effective and benefiting from parallelization, making it a flexible option for practical retrieval-based applications.

\vspace{-0.2cm}
\paragraph{Search stage.}
The search stage processes online incoming queries and retrieves relevant items from the indexed data. As shown in the table, \oursb achieves significantly higher efficiency compared to BM25 and performs on par with dense retrieval methods when utilizing GPU resources. This advantage arises because the BoT index $V_{\text{BoT}}(\mathcal{D})$ has a fixed dimensionality, enabling tensorization for inner product calculations on the GPU. In contrast, BM25 term-based index $V_{\text{BM25}}(\mathcal{D})$ operates with millions of dimensions and relies on an inverted index for efficiency. 

\vspace{-0.2cm}
\section{Analysis}
% \subsection{Ablation Study}
We assess the impact of proposed components and various influencing factors, as detailed in Table~\ref{table:abl}.

\vspace{-0.2cm}
\begin{wraptable}{r}{6cm}
\centering
\vspace{-0.5cm}
\caption{Ablation study of \oursf and \oursb on NQ dataset.}
\scalebox{0.7}{
\begin{tabular}{lccc}
\hline
 & top1 & top5 & top20 \\ \hline

\multicolumn{4}{l}{\textit{Parametric Index}} \\
\quad \oursf & \textbf{49.1} & \textbf{69.3} & \textbf{80.7} \\
\quad w/ retrieved neg ($m$=1) & 47.9 & 68.4 & 79.6 \\
\quad w/ retrieved neg ($m$=100) & 47.3 & 69.0 & 80.5 \\
\quad w/ retrieved neg (MARCO) & 39.7 & 63.9 & 77.5 \\
\quad w/ retrieved neg (WIKI 8m) & 48.3 & 69.1 & 80.6 \\
\quad w/o retrieved neg & 44.9 & 66.9 & 78.8 \\
\quad w/o neg & 30.2 & 57.4 & 75.1 \\
\quad w/o SP objective & 43.8 & 68.0 & 79.9 \\ \hline
\multicolumn{4}{l}{\textit{Non-parametric Index}} \\
\quad \oursb & 39.8 & \textbf{62.9} & \textbf{76.3} \\
\quad w/ retrieved neg ($m$=1) & \textbf{41.2} & 62.3 & 76.4 \\
\quad w/ retrieved neg ($m$=100) & 37.3 & 62.4 & 76.5 \\
\quad w/ retrieved neg (MARCO) & 29.5 & 54.8 & 70.1 \\
\quad w/ retrieved neg (WIKI 8m) & 37.5 & 62.4 & 76.3 \\
\quad w/o retrieved neg & 32.3 & 56.0 & 72.1 \\
\quad w/o neg & 24.4 & 49.1 & 68.2 \\
\quad w/o SP objective & 12.3 & 30.0 & 46.8 \\
\quad w/ vary lentgh & 37.5 & 61.2 & 76.1 \\ \hline
\end{tabular}
}
\label{table:abl}
\vspace{-0.5cm}
\end{wraptable}

\paragraph{Impact of proposed components.}
Our ablation study confirms the significance of each component in our approach. Removing the semi-parametric loss (w/o SP objective) leads to a drop in accuracy of 5.3\% for \oursf and a substantial 31.5\% for \oursb, rendering beta search non-functional. Moreover, excluding in-training retrieved negatives (w/o retrieved neg) results in a decrease in top-1 accuracy: 4.2\% for \oursf and 7.5\% for \oursb. These results highlight the effectiveness of using beta search for in-training retrieval. Unlike static BM25 negatives, which quickly diminish in effectiveness as the model learns, beta search utilizes parametric queries that evolve with the model, continually ensuring that the negatives are challenging and relevant throughout the training process.

\vspace{-0.2cm}
\paragraph{Effect of negative sample hardness.}
We explore how the difficulty of retrieved negatives affects model effectiveness. The parameter $m$ indicates the size of the passage pool from which negatives are identified and then randomly drawn, with lower $m$ values yielding harder negatives. While these harder negatives can improve contrastive learning, they also increase the risk of misclassifying weak positives as negatives. Our results (w/ retrieved neg m=\{1,100\}) indicate that adjusting $m$ to 1 or 100, compared to the baseline of 20, degrades the performance. Thus, an 
$m$ value of 20 provides the optimal balance, effectively challenging the model while reducing the likelihood of misclassification.

\vspace{-0.2cm}
\paragraph{Effect of negative sample source.}
Our results (w/ retrieved neg MARCO) indicate that switching the source of negative samples from the Wikipedia corpus to the MS MARCO with 8.8 million passages leads to a notable drop in performance. In a parallel experiment (w/ retrieved neg WIKI 8m) using a Wikipedia corpus of the same size, performance remained consistent with our baseline, indicating that corpus size is not the main factor behind the observed decline. Instead, the source of negatives plays a crucial role in performance. The disparity stems from differences in the writing styles and structures unique to each corpus (e.g., Wikipedia passages typically include a short title preceding the text), which cause the model to focus on superficial, corpus-specific features rather than developing a deeper understanding of the relevance.

\vspace{-0.2cm}
\paragraph{Impact of text length on non-parametric index.}
Unlike the BM25 term-based index, the BoT index lacks term weighting, meaning longer texts may activate more dimensions, resulting in higher inner product scores. To assess the impact of text length on the effectiveness of \oursb, we re-segmented the Wikipedia corpus into passages ranging from 50 to 200 words, while maintaining the same overall number of passages. Our results (w/ vary lentgh) show that the top-1 accuracy of \oursb decreased slightly from 39.8\% to 37.5\%, indicating minimal impact on performance. This slight drop can be explained by the sub-linear growth in unique tokens as text length increases and the high sparsity of the representations, where increasing activations has little impact on relevance.

\begin{wraptable}{r}{6cm}
\vspace{-0.8cm}
\centering
\caption{In-training retrieval latency per batch, with the storage size and GPU memory allocation for corresponding index.}
\scalebox{0.95}{
\begin{tabular}{l|ccc}
\hline
Method                           & Latency       & Storage & GPU \\ \hline
BM25                             & 3s            & 2.3GB   & /              \\
DPR                              & <1ms          & 31.5GB  & 31GB           \\
\oursb & <1ms          & 2.7GB   & 10GB           \\ \hline
\end{tabular}
}
\label{table:in-training-ret}
\vspace{-0.25cm}
\end{wraptable}

\vspace{-0.2cm}
\paragraph{Comparison of in-training retrieval.}
We compared in-training retrieval across different systems, as shown in Table~\ref{table:in-training-ret}. Compared to BM25, \oursb has up to 30x lower latency when using GPU resources for large corpora. In contrast to dense retrieval methods like \text{DPR}, \oursb requires less GPU allocation and uses a fixed index, which eliminates the need for periodic re-indexing during the training loop and ensures that the training objective is not compromised by a stale index. Additional details can be found in Appendix~\ref{sec:appendix-in-training-ret}.

%\vspace{-0.2cm}
\section{Related Work}
%\vspace{-0.2cm}
\paragraph{Disentangled Retrieval} 
Disentangled retrieval, also known as sparse lexical retrieval, develops sparse representations for queries and documents within a pre-defined vocabulary space, where each dimension reflecting the importance of a specific token. These methods have proven effective in text matching~\citep{dai2020context,bai2020sparterm,formal2021splade,formal2022distillation,ram2022you} and have been utilized to enhance search efficiency in subsequent studies~\citep{gao2021coil,shen2022lexmae,lin2023aggretriever,lin2023dense}. Notably, several works like TILDE~\citep{zhuang2021tilde,zhuang2021fast}, and SPARTA~\citep{zhao2021sparta} use bag-of-tokens query representations for efficient online query processing. These methods fall under the category of semi-parametric retrieval as they employ non-parametric representations on the query side. Complementing these efforts, our work focuses on addressing the challenges associated with the index side, which is inherently more complex due to the greater length and contextual depth of documents. 
We discuss the taxonomy of neural retrieval in more detail in Appendix~\ref{sec:appendix-texonomy}.

%\vspace{-0.2cm}
\paragraph{In-training Retrieval} 
Retrieving data in the training loop of retrieval models is an emerging yet challenging practice that serves several critical purposes. 
This includes acquiring negative samples for contrastive learning~\citep{zhan2021optimizing,robinson2021contrastive}, sourcing relevant instances for data augmentation~\citep{blattmann2022retrieval,shi2023replug}, and facilitating the training of retrieval-based language models~\citep{asai2023retrieval} in an end-to-end manner. 
However, this process is complicated due to the need for frequent re-indexing of the corpus as the training of the retriever progresses. 
Recent research has explored strategies like asynchronous index updates~\citep{guu2020retrieval,xiong2020approximate,izacard2022few,shi2023replug} or building temporary indexes on-the-fly from the current training batch~\citep{zhong2022training,min2022nonparametric}.
Our work proposes a semi-parametric framework which supports a non-parametric index, thereby avoiding these complications and streamlining the in-training retrieval practice.

\vspace{-0.3cm}
\section{Conclusions}
\label{sec:conclusion}
\vspace{-0.1cm}
In this paper, we introduce \ours, a semi-parametric bi-encoder retrieval framework that supports both parametric and non-parametric indexes to address the emerging needs of retrieval-based applications. Unlike traditional neural retrieval methods that rely solely on embeddings as indexes, \ours additionally incorporates a non-parametric bag-of-tokens index. The flexibility of \ours makes it particularly well-suited for applications requiring efficient or low-cost indexing and facilitates co-training with a fixed index.

\subsubsection*{Acknowledgments}
Lei Chen’s work is partially supported by National Key Research and Development Program of China Grant No. 2023YFF0725100, National Science Foundation of China (NSFC) under Grant No. U22B2060, Guangdong-Hong Kong Technology Innovation Joint Funding Scheme Project No. 2024A0505040012, the Hong Kong RGC GRF Project 16213620, RIF Project R6020-19, AOE Project AoE/E-603/18, Theme-based project TRS T41-603/20R, CRF Project C2004-21G, Guangdong Province Science and Technology Plan Project 2023A0505030011, Guangzhou municipality big data intelligence key lab, 2023A03J0012, Hong Kong ITC ITF grants MHX/078/21 and PRP/004/22FX, Zhujiang scholar program 2021JC02X170, Microsoft Research Asia Collaborative Research Grant, HKUST-Webank joint research lab and 2023 HKUST Shenzhen-Hong Kong Collaborative Innovation Institute Green Sustainability Special Fund, from Shui On Xintiandi and the InnoSpace GBA.

\bibliography{iclr2025_conference}
\bibliographystyle{iclr2025_conference}

\appendix
\section{Taxonomy of Neural Retrieval}
\label{sec:appendix-texonomy}
In this section, we outline the taxonomy of existing neural retrievers, discussing distinctions such as disentangled versus entangled, dense versus sparse, and parametric versus semi-parametric. This classification aims to clarify the concepts discussed throughout our paper.

\textbf{Entangled vs. Disentangled Retriever} differ in the type of representations they use for search:
\begin{itemize}[noitemsep, topsep=0pt, left=0pt]
\setlength{\itemsep}{0pt}
\item \textbf{Entangled retriever} typically employs latent representations with dimensions such as 512, 768, 1024, or 2048 for inner-product search. These representations can be learned flexibly and effectively without prior knowledge, as they are entangled. An entangled retriever can also use sparse latent representations, such as the BPR method~\citep{yamada2021efficient}, which employs a learned hash function on a 768-dimensional latent vector for efficient searching.

\item \textbf{Disentangled retriever}, on the other hand, utilizes disentangled representations with much larger dimensionality, spanning from tens of thousands, representing the language model's vocabulary, to millions, typical of BM25 vocabulary. Each dimension corresponds to the importance of a specific token within the vocabulary. Disentangled retrieval methods rely on disentangled representations as a prior, which leads to desirable properties such as interpretability, controllability, robustness, and, as discussed in this paper, accelerated indexing. Previous work~\citep{zhou2024retrievalbased, formal2021splade_v2, shen2022lexmae} has also suggested that using disentangled representations improves embedding quality, as foundational models are optimized in this disentangled representation space to predict mask tokens or next tokens, rather than in the entangled representation space. Note that disentangled retrievers can also use dense (fully activated) representations. For example, \text{VDR}~\citep{zhou2024retrievalbased} utilizes disentangled representations that can be fully activated for search.
\end{itemize}

Thus, the classification of entangled vs. disentangled is independent of whether the representation is dense or sparse. Generally, entangled retrieval methods tend to use dense representations, while disentangled retrieval methods often use sparse ones, but this is not a strict requirement.

\textbf{Dense vs. Sparse Retriever} differ based on whether their representations are fully activated or have been sparsified. Typically, entangled representations are fully activated, while disentangled representations are often sparsified to reduce storage size and to facilitate the construction of an inverted index for efficient searching. Consequently, dense retrieval is commonly associated with entangled representations and sparse retrieval with disentangled ones. However, the relationship between entangled/disentangled and dense/sparse retrieval is not rigid. Exceptions exist where entangled representations can be sparsified~\citep{yamada2021efficient} and disentangled representations can be fully activated~\citep{zhou2024retrievalbased}. 

\paragraph{Full Parametric vs. Semi-parametric Retrieval} 
These approaches differ based on the utilization of neural parameters for encoders. Full parametric retrieval systems employ neural parameters for both encoders. In contrast, semi-parametric retrieval systems use one neural encoder alongside one non-parametric encoder, typically involving tokenization-based representations. To our knowledge, existing semi-parametric systems predominantly engage in disentangled retrieval, as they all utilize tokenization-based non-parametric representations that operate on a vocabulary space. Notable examples include TILDE~\citep{zhuang2021tilde}, TILDE-v2~\citep{zhuang2021fast}, and SPARTA~\citep{zhao2021sparta}, which implement tokenization-based representations on the query side for efficient online query processing. Our work, \ours, represents a unique approach within this category, offering semi-parametric retrieval that utilizes binary bag-of-tokens representations on the index side for emerging scenarios.
\section{Details of Latency Evaluation}
\label{sec:appendix-efficiency}
During the indexing stage, $E_{\theta}$ and $T$ operate with a batch size of 32 and a maximum text length of 256. In the search stage, the computation of $\text{f}(q, \mathcal{D})$ includes both inner product calculation and sorting of the top-$k$ passages. Queries are processed in batches of 32, with passage embeddings stored in either CPU or GPU memory using half-precision floating-point (FP16) to optimize memory usage. Our analysis excludes the time spent on I/O and data type conversion between CPU and GPU, assuming sufficient processing resources are available.

For BM25, we utilize Pyserini~\citep{lin2021pyserini}, a library based on a Java implementation developed around Lucene. For neural retrieval, our implementation is in Python, leveraging PyTorch's sparse module\footnote{\url{https://pytorch.org/docs/stable/sparse.html}} for efficient inner product computation, without building an inverted index. Timing measurements are performed by running each operation 10 times and reporting the average after excluding the maximum and minimum values. To avoid out-of-memory issues on our devices, we perform searches in batches on a 1-million document corpus and accumulate the latencies. Note that inverted indexes rely heavily on memory access and integer operations, which are generally inefficient on GPU architectures.
\section{In-training Retrieval Details}
\label{sec:appendix-in-training-ret}

\begin{wraptable}{t}{6cm}
\vspace{-1.2cm}
\centering
\scalebox{0.7}{
\begin{tabular}{lccc}
\hline
 & Latency (ms) & Storage (GB) & GPU (GB) \\ \hline
 & \multicolumn{3}{c}{Index Density} \\ \cline{2-4} 
a=256 & 0.20 & 2.8 & 6.9 \\
a=512 & 0.21 & 4.9 & 23.5 \\
a=1024 & 0.21 & 9.0 & 46.7 \\ \hline
 & \multicolumn{3}{c}{Query Batch Size} \\ \cline{2-4} 
bs=32 & 0.20 & / & / \\
bs=128 & 0.21 & / & / \\
bs=512 & 0.24 & / & / \\ \hline
\end{tabular}
}
\caption{Retrieval latency, index storage size, and GPU allocation for \oursb.}
\label{table:appendix-ret}
\vspace{-0.2cm}
\end{wraptable}

We conducted a simulation test to evaluate factors affecting in-training retrieval, summarized in Table~\ref{table:appendix-ret}. Initially, we built a binary token index with dimensions of 30k and a sample size of 500m, where each vector consists of 256 dimensional activated. We then varied the density of passage representations by adjusting the activation number from 256 to 512 and 1024. As the activation number increased, storage and GPU allocation also increased, while latency remained largely unchanged. Additionally, we found query batch size had minimal impact on latency.

\section{Analysis on Term Weighting and Expansion}
\label{sec:term}

To systematically compare our method with BM25, we control for vocabulary differences by using BM25 with the same BERT-base-uncased vocabulary as ours. This ensures both methods share the same dimensionality for sparse representation, differing only in two key aspects:
\begin{itemize}
    \setlength{\itemindent}{-2pt}
    \item Term expansion: BM25 uses only lexical tokens, while \text{SiDR} allows for term expansion.
    \item Term weighting: BM25 relies on statistical-based term weights, whereas \text{SiDR} learns contextualized weights.
\end{itemize}
In this section, we empirically demonstrate how these factors -- term expansion and term weighting -- affect the outcome. Below, we introduce two additional representation forms:

\paragraph{Lexical Parametric Representation $V_{\theta}^{lex}(x)$} This representation activates only the lexical tokens present in the input text $x$. It leverages learned term weights but does not permit term expansion:
\begin{equation} 
\begin{split} 
    V_{\theta}^{lex}(x) = V_{\text{BoT}} \circ V_{\theta}(x)  \nonumber
\end{split} 
\end{equation}
\paragraph{Binary Parametric Representation $V_{\theta}^{bin}(x)$} This representation activates the same number of tokens but assigns uniform weights (set to one), removing learned scalar term weighting. It allows term expansion but does not apply term weights:
\begin{equation}
\begin{split}
  V_{\theta}^{bin}(x) = \text{Binarize} \circ V_{\theta}(x)  \nonumber
\end{split}
\end{equation}
where \text{Binarize} is a function mapping non-zero values to one.

To independently assess the impact of term expansion and term weighting, we propose several variants of $\text{SiDR}_\text{full}$: $\text{SiDR}_\text{full}$ (w/o weight at doc) and $\text{SiDR}_\text{full}$ (w/o expand at doc), which utilize $V_{\theta}^{bin}(p)$ and $V_{\theta}^{lex}(p)$ on the document side during inference.
Additionally, we introduce variants of $\text{SiDR}_\beta$: $\text{SiDR}_\beta$ (w/o weight at query) and $\text{SiDR}_\beta$ (w/o expand at query), which employ $V_{\theta}^{bin}(q)$ and $V_{\theta}^{lex}(q)$ on the query side at inference.
Furthermore, we propose $\text{SiDR}_{full}$ (w/o expand at doc, training), which is trained with $V_{\theta}(p)$ replaced by $V_{\theta}^{\text{lex}}(p)$ to ensure consistency between training and inference phases. All these models are compared against BM25 that utilizes the same bert-base-uncased tokenization and vocabulary. This controls for vocabulary differences, isolating the effects of term selection and term weighting. We also include an extreme baseline that uses a bag-of-tokens representation for both the query and the passage, referred to as BoT overlap.

\begin{table*}[ht]
  \centering
  \caption{Ablation study on learned term weighting and term expansion, with results reported as top-1 accuracy on NQ test splits.}
  \scalebox{0.8}{
% Table generated by Excel2LaTeX from sheet 'exp-index'
\begin{tabular}{lccccc}
\toprule
\multirow{2}[4]{*}{Model} & \multicolumn{2}{c}{Query} & \multicolumn{2}{c}{Document} & \multirow{2}[4]{*}{Accuracy} \\
\cmidrule{2-5}   & Expand & Weight & Expand & Weight &  \\
\midrule
BM25 (bert-base-uncased) & $\times$ & $\checkmark$ & $\times$ & $\checkmark$ & 21.9 \\
\midrule
\multicolumn{6}{l}{\textit{\textbf{Ablation of $\text{SiDR}_{\text{full}}$ on doc side}}} \\
\quad $\text{SiDR}_{\text{full}}$ & \multirow{5}[1]{*}{$\checkmark$} & \multirow{5}[1]{*}{$\checkmark$} & $\checkmark$ & $\checkmark$ & 49.1 \\
\quad $\text{SiDR}_{\text{full}}$ (w/o weight at doc) &    &    & $\checkmark$ & $\times$ & 33.1 \\
\quad $\text{SiDR}_{\text{full}}$ (w/o expand at doc) &    &    & $\times$ & $\checkmark$ & 38.9 \\
\quad $\text{SiDR}_{\text{full}}$ (w/o expand at doc, training) &    &    & $\times$ & $\checkmark$ & 43.1 \\
\quad $\text{SiDR}_{\text{beta}}$ &    &    & $\times$ & $\times$ & 39.8 \\
\midrule
\multicolumn{6}{l}{\textit{\textbf{Ablation of $\text{SiDR}_{\text{beta}}$ on query side}}} \\
\quad $\text{SiDR}_{\text{beta}}$  (w/o weight at query) & $\checkmark$ & $\times$ & \multirow{3}[1]{*}{$\times$} & \multirow{3}[1]{*}{$\times$} & 14.5 \\
\quad $\text{SiDR}_{\text{beta}}$  (w/o expand at query) & $\times$ & $\checkmark$ &    &    & 34.3 \\
\quad BoT overlap & $\times$ & $\times$ &    &    & 14.2 \\
\bottomrule
\end{tabular}%
}
\label{tab:abl1}
\end{table*}

When using a parametric index, we conduct an ablation study on $\text{SiDR}_\text{full}$ to assess the impact of removing term weighting and term expansion on the document side. From top to bottom, we systematically remove term weight or expansion, simplifying the index of $\text{SiDR}_\text{full}$ to assess their individual contributions. Our results indicate that removing either term weight or term expansion leads to worse outcomes than removing both (i.e., $\text{SiDR}_{\beta}$). This is because our training objective is specifically designed to align query embeddings with the BoT index, rather than these variations. Furthermore, we find that if the training is adjusted to accommodate these variations, such as document representations without term expansion, these variations can outperform the BoT index. This demonstrates that neural bi-encoders have great learning potential, with improvement stemming from not only the training itself but also how well the training aligns with inference.

When using a bag-of-tokens index, we assess the impact of term weight and expansion on the query side. Starting from a baseline uses unweighted term overlap, referred to as ``BoT overlap'', applying BM25's term weights to both queries and documents yields a 7.7\% improvement. In comparison, our method's learned query term weights achieve a 20.1\% improvement, while learned term expansion provides minimal additional gain. Combining term weights and expansion on the query side results in a 25.6\% improvement, which is $\text{SiDR}_{\beta}$.

In conclusion, when using an embedding index, we demonstrate that both learned term weighting and term expansion on the document side are crucial. Conversely, when using a bag-of-tokens index, the improvements primarily come from term weighting on the query side rather than expansion. Furthermore, ensuring consistency between training and inference representations is essential. A parametric index can underperform compared to non-parametric ones if the representations used during inference do not align with those used during training.

\section{Cost-Effectiveness Analysis}
\label{sec:late-para}

\begin{table*}[ht]
  \centering
  \caption{Additional late-parametric baselines and cost-effectiveness analysis performed on a retrieval task using 3.6k NQ test queries across a 21 million Wikipedia corpus. Costs are determined by the number of text chunks (including both queries and passages) that require embedding by neural encoders. Parentheses indicate the ratio of text chunks embedded to the total retrieval corpus.}
  \scalebox{0.8}{

\begin{tabular}{lcc}
\toprule
   & Performance & Cost \\
\midrule
\multicolumn{3}{l}{\textit{\textbf{Non-parametric Index}}} \\
\quad BM25 & 22.7 & 0 \\
\quad \oursb & 39.8 & 3.6k (0.01\%) \\
\midrule
\multicolumn{3}{l}{\textit{\textbf{Late-parametric with top-100 rerank}}} \\
\quad BM25 + VDR & 41.6 & 364k (1.73\%) \\
\quad BM25 + SiDR & 44.0 & 364k (1.73\%) \\
\quad BM25 + Contriever & 39.3 & 364k (1.73\%) \\
\quad BM25 + $\text{E5}_\text{base}$ & 50.4 & 364k (1.73\%) \\
\quad $\text{SiDR}_{\beta}$ (m=100) & 50.3 & 364k (1.73\%) \\
\quad \oursb + VDR & 43.2 & 367k  (1.74\%) \\
\quad \oursb + Contriever & 42.9 & 367k  (1.74\%) \\
\quad \oursb + $\text{E5}_\text{base}$ & 57.7 & 367k  (1.74\%) \\
\midrule
\multicolumn{3}{l}{\textit{\textbf{Parametric Index}}} \\
\quad \oursf & 49.1 & 21m (100.01\%) \\
\quad Contriever & 41.5 & 21m (100.01\%) \\
\quad $\text{E5}_\text{base}$ & 57.9 & 21m (100.01\%) \\
\bottomrule
\end{tabular}%
}
\label{tab:late-para}
\end{table*}

Late parametric retrieval aims to provide a quick-start and low-cost search initialization through a non-parametric index, while simultaneously building a parametric index during the search service, eventually transitioning to a fully parametric index for searching. To fulfill this requirement, the first-stage utilizes a retriever that supports a non-parametric index, while the second-stage retriever can be any parametric bi-encoder. This method can be seen as a subset of hybrid retrieval systems~\citep{leonhardt2022efficient,gao2021complement}, with specific choices constrained to the two stages.

We introduce various combinations of BM25 and $\text{SiDR}_{\beta}$ as the first-stage retriever, paired with more advanced retrievers in the second stage to demonstrate their effectiveness. The results of these combinations are presented in Table~\ref{tab:late-para}. 
Moreover, we assess the cost-effectiveness of these frameworks, particularly in scenarios where raw data has not been indexed. In such cases, the primary cost arises from the neural model's forward pass for text embedding. Therefore, we measure cost by counting the number of text chunks (both query and passage) that require embedding.

For BM25, no neural embedding is required. While $\text{SiDR}_{\beta}$ employs a bag-of-tokens index, waiving the indexing cost, it requires embedding 3.6k queries (0.01\% of the corpus) to complete the retrieval task. Despite this, it offers a significant performance improvement of 17.1\% in accuracy over BM25. 
For various late parametric baselines, an additional embedding of 100 passages per query is needed --- approximately 1.7\% of the corpus --- yet this results in further improvements over BM25. Conversely, the conventional retrieval pipeline, which requires embedding the entire corpus to achieve performance comparable to that of late parametric models with top-100 reranking. This analysis shows that semi-parametric models provide a more cost-effective solution by balancing retrieval performance with computational efficiency.

Among various late-parametric baselines, $\text{SiDR}_{\beta}$ consistently outperforms BM25 as the first-stage retriever. However, this advantage requires embedding an additional 3.6k queries if other models are employed as the second-stage retriever. In exploring second-stage retrievers, we have tested state-of-the-art models like E5 and Contriever. Our results indicate that stronger retrievers lead to better overall late-parametric performance. Notably, in all our tests, $\text{SiDR}_{\beta}$ combined with any second-stage model consistently outperforms BM25 paired with the same model. Furthermore, when $\text{SiDR}_{\beta}$ serves as the first-stage retriever and re-ranks the top-100 passages, its performance is comparable to, and often exceeds, that of full parametric search with these retrievers.

\end{document}